\def\year{2020}\relax
\documentclass[letterpaper]{article} 
\usepackage{aaai20}  
\usepackage{times}  
\usepackage{helvet} 
\usepackage{courier}  
\usepackage[hyphens]{url}  
\usepackage{graphicx} 
\usepackage{graphicx}  
\usepackage{amssymb,amsmath}
\usepackage{booktabs}
\usepackage{multirow}
\usepackage{array}

\newcommand{\citet}[1]{\citeauthor{#1} \shortcite{#1}}
\newcommand{\citep}{\cite}

\title{Multi-channel Reverse Dictionary Model}

\author{
Lei Zhang$^{2}\thanks{Indicates equal contribution}\thanks{Work done during internship at Tsinghua University}$,
Fanchao Qi$^{12*}$, 
Zhiyuan Liu$^{12}\thanks{Corresponding author}$,
{\Large {\bf Yasheng Wang$^{3}$,
Qun Liu$^{3}$,
Maosong Sun$^{12}$ 
}}\\ 
$^{1}$Department of Computer Science and Technology, Tsinghua University \\
$^{2}$Institute for Artificial Intelligence, Tsinghua University \\
Beijing National Research Center for Information Science and Technology\\
$^{3}$Huawei Noah's Ark Lab\\
{ zhanglei9003@gmail.com, qfc17@mails.tsinghua.edu.cn}\\
{ \{liuzy, sms\}@tsinghua.edu.cn, \{wangyasheng, qun.liu\}@huawei.com}
}

\begin{document}
\maketitle

\begin{abstract}
A reverse dictionary takes the description of a target word as input and outputs the target word together with other words that match the description. 
Existing reverse dictionary methods cannot deal with highly variable input queries and low-frequency target words successfully. 
Inspired by the description-to-word inference process of humans, we propose the multi-channel reverse dictionary model, which can mitigate the two problems simultaneously.
Our model comprises a sentence encoder and multiple predictors.
The predictors are expected to identify different characteristics of the target word from the input query. 
We evaluate our model on English and Chinese datasets including both dictionary definitions and human-written descriptions.
Experimental results show that our model achieves the state-of-the-art performance, and even outperforms the most popular commercial reverse dictionary system on the human-written description dataset. 
We also conduct quantitative analyses and a case study to demonstrate the effectiveness and robustness of our model.
All the code and data of this work can be obtained on \url{https://github.com/thunlp/MultiRD}.
\end{abstract}

\section{Introduction}
\begin{figure}[!t]
    \centering
    \includegraphics[width=.95\columnwidth]{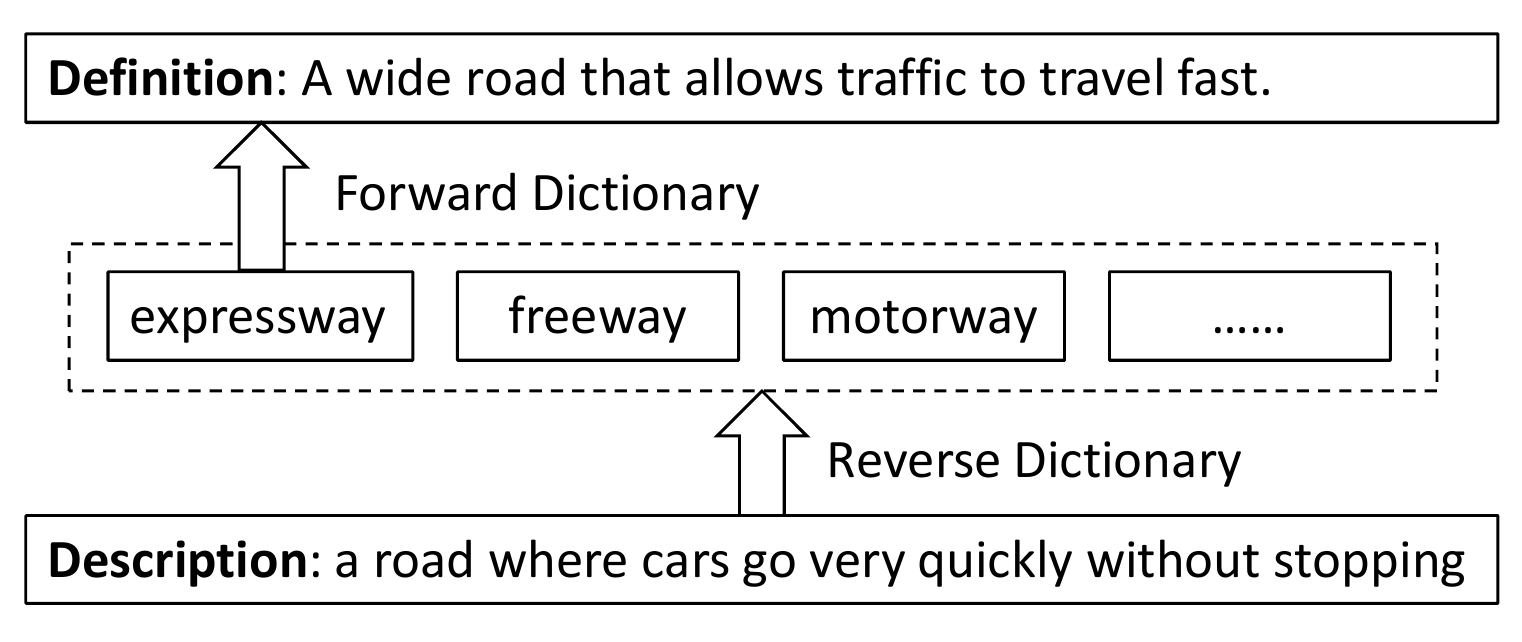}
    \caption{An example illustrating what a forward and a reverse dictionary are.}
    \label{fig:reverse-dictionary-example}
\end{figure}

A regular (forward) dictionary maps words to definitions while a \textit{reverse dictionary} \citep{sierra2000onomasiological} does the opposite and maps descriptions to corresponding words. 
In Figure \ref{fig:reverse-dictionary-example}, for example, a regular dictionary tells you that ``expressway'' is ``a wide road that allows traffic to travel fast'', and when you input ``a road where cars go very quickly without stopping'' to a reverse dictionary, it might return ``expressway'' together with other semantically similar words like ``freeway''.

Reverse dictionaries have great practical value.
First and foremost, they can effectively address the tip-of-the-tongue problem \citep{brown1966tip}, which severely afflicts many people, especially those who write a lot such as researchers, writers and students. 
Additionally, reverse dictionaries can render assistance to new language learners who know a limited number of words.
Moreover, reverse dictionaries are believed to be helpful to word selection (or word dictionary) anomia patients, people who can recognize and describe an object but fail to name the object due to neurological disorder \citep{benson1979neurologic}.
In terms of natural language processing (NLP), reverse dictionaries can be used to evaluate the quality of sentence representations \citep{Hill2016LearningTU}.
They are also beneficial to the tasks involving text-to-entity mapping including question answering and information retrieval \citep{kartsaklis2018mapping}.

There have been some successful commercial reverse dictionary systems such as OneLook\footnote{\url{https://onelook.com/thesaurus/}}, the most popular one, but their architecture is usually undisclosed proprietary knowledge. 
Some scientific researches into building reverse dictionaries have also been conducted. 
Early work adopts sentence matching based methods, which utilize hand-engineered features to find the words whose stored definitions are most similar to the input query \citep{bilac2004dictionary,zock2004word,Mndez2013ARD,Shaw2013BuildingAS}.
But these methods cannot successfully cope with the main difficulty of reverse dictionaries that human-written input queries might differ widely from target words' definitions.

\citet{Hill2016LearningTU} propose a new method based on neural language model (NLM). 
They employ a NLM as the sentence encoder to learn the representation of the input query, and return those words whose embeddings are closest to the input query's representation.
The NLM based reverse dictionary model alleviates the above-mentioned problem of variable input queries, but its performance is heavily dependent on the quality of word embeddings. 
According to Zipf's law \citep{zipf1949human}, however, quite a few words are low-frequency and usually have poor embeddings, which 
will undermine the overall performance of ordinary NLM based models.



To tackle the issue, we propose the \textbf{multi-channel reverse dictionary model}, which is inspired by the description-to-word inference process of humans.
Taking ``expressway'' as an example, when we forget what word means ``a road where cars go very quickly'', it may occur to us that the \textit{part-of-speech tag} of the target word should be ``noun'' and it belongs to the \textit{category} of  ``entity''. 
We might also guess that the target word probably contains the \textit{morpheme} ``way''.
When having knowledge of these characteristics, it is much easier for us to search the target word out. 
Correspondingly, in our multi-channel reverse dictionary model, we employ multiple predictors to identify different characteristics of target words from input queries. 
By doing this, the target words with poor embeddings can still be picked out by their characteristics and, moreover, 
the words which have close embeddings to the correct target word but contradictory characteristics to the given description will be filtered out.

We view each characteristic predictor as an information channel of searching the target word. 
Two types of channels involving internal and external channels are taken into consideration.
The internal channels correspond to the characteristics of words themselves including the part-of-speech (POS) tag and morpheme.
The external channels reflect characteristics of target words related to external knowledge bases.
We take account of two external characteristics including the word category and \textit{sememe}.
The word category information can be obtained from word taxonomy systems and it usually corresponds to the genus words of definitions.
A sememe is defined as the minimum semantic unit of human languages \citep{bloomfield1926set}, which is similar to the concept of \textit{semantic primitive} \citep{wierzbicka1996semantics}.
Sememes of a word depict the meaning of the word atomically, which can be also predicted from the description of the word.

More specifically, we adopt the well-established bi-directional LSTM (BiLSTM) \citep{hochreiter1997long} with attention \citep{Bahdanau2015NeuralMT} as the basic framework and add four feature-specific characteristic predictors to it.
In experiments, we evaluate our model on English and Chinese datasets including both dictionary definitions and human-written descriptions, finding that our model achieves the state-of-the-art performance. 
It is especially worth mentioning that for the first time OneLook is outperformed when input queries are human-written descriptions. 
In addition, to test our model under other real application scenarios like crossword game, we provide our model with prior knowledge about the target word such as the initial letter, and find it yields substantial performance enhancement. 
We also conduct detailed quantitative analyses 
and a case study to demonstrate the effectiveness of our model as well as its robustness in handling polysemous and low-frequency words.

\section{Related Work}
\subsection{Reverse Dictionary Models}

Most of existing reverse dictionary models are based on sentence-sentence matching methods, i.e., comparing the input query with stored word definitions and return the word whose definition is most similar to the input query \citep{zock2004word,bilac2004dictionary}.
They usually use some hand-engineered features, e.g., \textit{tf-idf}, to measure sentence similarity, and leverage well-established information retrieval techniques to search the target word \citep{Shaw2013BuildingAS}.
Some of them utilize external knowledge bases like WordNet \citep{Miller1995WordNet} to enhance sentence similarity measurement by finding synonyms or other pairs of related words between the input query and stored definitions \citep{Mndez2013ARD,Lam2013CreatingRB,Shaw2013BuildingAS}. 

Recent years have witnessed a growing number of reverse dictionary models which conduct sentence-word matching.
\citet{Thorat2016ImplementingAR} present a node-graph architecture which can directly measure the similarity between the input query and any word in a word graph. 
However, it works on a small lexicon ($3,000$ words) only. 
\citet{Hill2016LearningTU} propose a NLM based reverse dictionary model, which uses a bag-of-words (BOW) model or an LSTM to embed the input query into the semantic space of word embeddings, and returns the words whose embeddings are closest to the representation of the input query.

Following the NLM model, 
\citet{MorinagaY18} incorporate category inference to eliminate irrelevant results and achieve better performance;
\citet{kartsaklis2018mapping} employ a graph of WordNet synsets and words in definitions to learn target word representations together with a multi-sense LSTM to encode input queries, and they claim to deliver state-of-the-art results;
\citet{hedderich2019using} use multi-sense embeddings when encoding the queries, aiming to improve sentence representations of input queries;
\citet{pilehvar2019importance} adopt sense embeddings to disambiguate senses of polysemous target words.

Our multi-channel model also uses a NLM to embed input queries. 
Compared with previous work, 
our model employs multiple predictors to identity characteristics of target words, which is consistent with the inference process of humans, and achieves significantly better performance. 


\subsection{Applications of Dictionary Definitions}
Dictionary definitions are handy resources for NLP research. 
Many studies utilize dictionary definitions to improve word embeddings \citep{Noraset2017DefinitionML,Tissier2017Dict2vecL,bahdanau2017learning,Bosc2018AutoEncodingDD,ScheepersKG18}.
In addition, dictionary definitions are utilized in various applications including word sense disambiguation \citep{luo2018incorporating}, knowledge representation learning \citep{xie2016representation}, reading comprehension \citep{long2017world} and knowledge graph generation \citep{silva2018building,prokhorov2019generating}.

\section{Methodology}
\begin{figure}[t]
    \centering
    \includegraphics[width=.95\columnwidth]{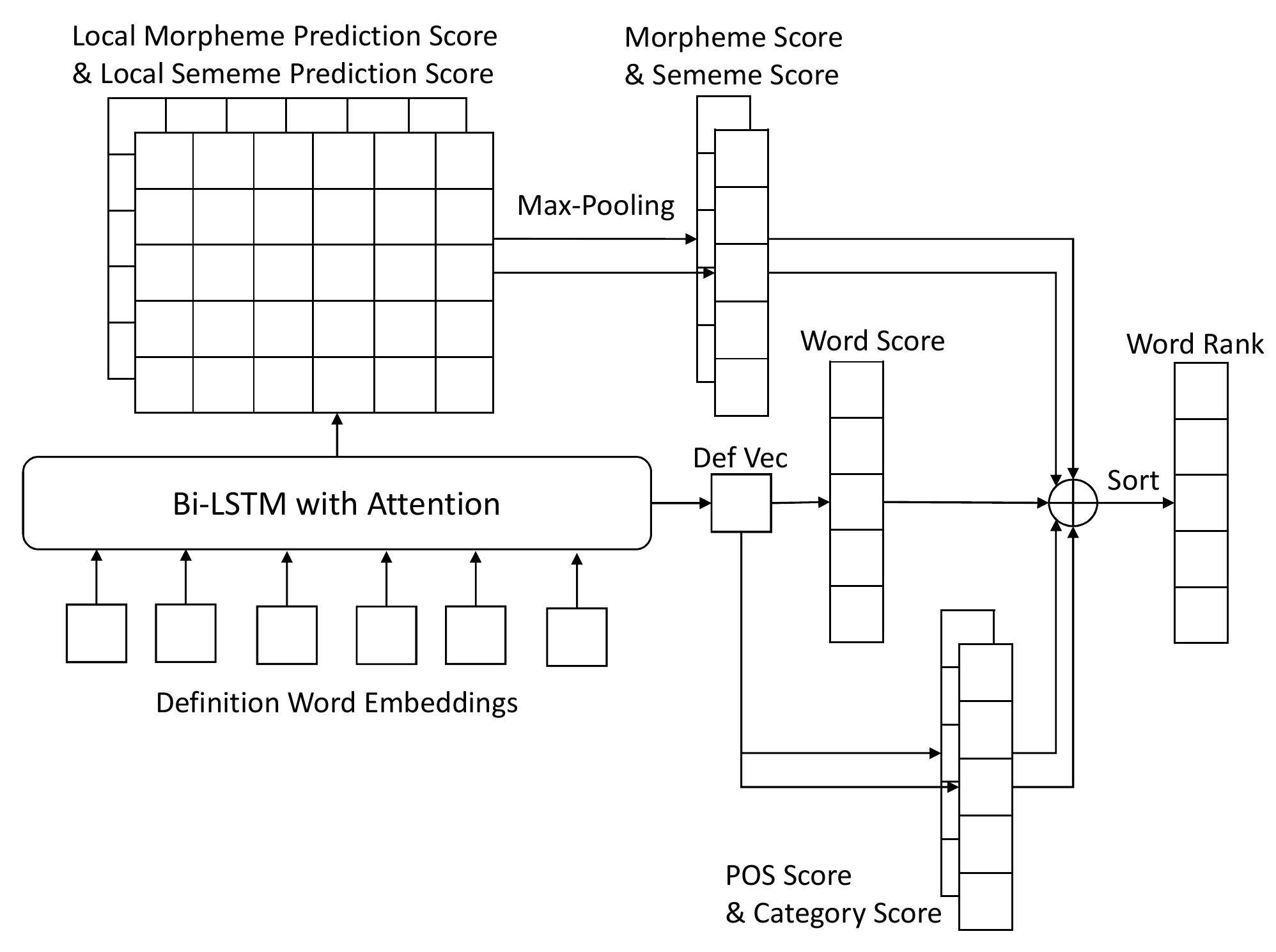}
    \caption{Multi-channel reverse dictionary model.}
    \label{fig:Multi-channel}
\end{figure}

In this section, we first introduce some notations. 
Then we describe our basic framework, i.e., BiLSTM with attention. 
Next we detail our multi-channel model and its two internal and two external predictors.
The architecture of our model is illustrated in Figure \ref{fig:Multi-channel}.

\subsection{Notations}
We define $\mathbb{W}$ as the vocabulary set, $\mathbb{M}$ as the whole morpheme set and $\mathbb{P}$ as the whole POS tag set.
For a given word $w\in \mathbb{W}$, its morpheme set is $\mathbb{M}_w = \{ m_1,\cdots,m_{|\mathbb{M}_w|} \}$, where each of its morpheme $m_i \in \mathbb{M}$ and $|\cdot|$ denotes the cardinality of a set.
A word may have multiple senses and each sense corresponds to a POS tag. 
Supposing $w$ has $n_w$ senses, all the POS tags of its senses form its POS tag set $\mathbb{P}_w=\{p_1,\cdots,p_{n_w}\}$, where each POS tag $p_i \in \mathbb{P}$.
In subsequent sections, we use lowercase boldface symbols to stand for vectors and uppercase boldface symbols for matrices. 
For instance, $\mathbf{w}$ is the word vector of $w$ and $\mathbf{W}$ is a weight matrix.




\subsection{Basic Framework}
The basic framework of our model is essentially similar to a sentence classification model, composed of a sentence encoder and a classifier. 
We select Bidirectional LSTM (BiLSTM) \citep{schuster1997bidirectional} as the sentence encoder, which encodes an input query into a vector.
Different words in a sentence have different importance to the representation of the sentence, e.g., the genus words are more important than the modifiers in a definition.   
Therefore, we integrate attention mechanism \citep{Bahdanau2015NeuralMT} into BiLSTM to learn better sentence representations.

Formally, for an input query $Q=\{q_1,\cdots,q_{|Q|}\}$, we first pass the pre-trained word embeddings of its words $\mathbf{q}_1,\cdots,\mathbf{q}_{|Q|} \in \mathbb{R}^d$ to the BiLSTM, where $d$ is the dimension of word embeddings, and obtain two sequences of directional hidden states:
\begin{equation}
    \begin{aligned}
        \{\stackrel{\rightarrow}{\mathbf{h}}_1,...,\stackrel{\rightarrow}{\mathbf{h}}_{|Q|}\},\{\stackrel{\leftarrow}{\mathbf{h}}_1,...,\stackrel{\leftarrow}{\mathbf{h}}_{|Q|}\}\\
        ={\rm BiLSTM}(\mathbf{q}_1,...,\mathbf{q}_{|Q|}),
    \end{aligned}
    \label{equation:bilstm}
\end{equation}
where $\stackrel{\rightarrow}{\mathbf{h}}_i, \stackrel{\leftarrow}{\mathbf{h}}_i \in \mathbb{R}^l$ and $l$ is the dimension of directional hidden states.
Then we concatenate bi-directional hidden states to obtain non-directional hidden states:
\begin{equation}
\begin{aligned}
    \mathbf{h}_i&={\rm Concatenate}(\stackrel{\rightarrow}{\mathbf{h}}_i,\stackrel{\leftarrow}{\mathbf{h}}_i). \\
\end{aligned}
\end{equation}
The final sentence representation is the weighted sum of non-directional hidden states:
\begin{equation}
\begin{aligned}
    \mathbf{v} &= \sum_{i=1}^{|Q|}\mathbf{\rm \alpha_i} {\mathbf{h}}_i,\\
\end{aligned}
\end{equation}
where $\alpha_i$ is the attention item serving as the weight:
\begin{equation}
\begin{aligned}
    \mathbf{\rm \alpha_i} &= \mathbf{h}_t \cdot \mathbf{h}_i,\\
    \mathbf{h}_t &= {\rm Concatenate}(\stackrel{\rightarrow}{\mathbf{h}}_{|Q|},\stackrel{\leftarrow}{\mathbf{h}}_1). \\
\end{aligned}
\end{equation}

Next we map $\mathbf{v}$, the sentence vector of the input query, into the space of word embeddings, and calculate the confidence score of each word using dot product:
\begin{equation}
\begin{aligned}
    \mathbf{v}_{word} &=\mathbf{W}_{word}   \mathbf{v}+\mathbf{b}_{word}, \\
    sc_{w,word} &= \mathbf{v}_{word} \cdot \mathbf{w},
\end{aligned}
\end{equation}
where ${sc}_{w, word}$ indicates the confidence score of $w$, $\mathbf{W}_{word} \in \mathbb{R}^{d \times 2l}$ is a weight matrix,  $\mathbf{b}_{word} \in \mathbb{R}^{d}$ is a bias vector.





\subsection{Internal Channel: POS Tag Predictor}


A dictionary definition or human-written description of a word is usually able to reflect the POS tag of the corresponding sense of the word.
We believe that predicting the POS tag of the target word can alleviate the problem of returning words with POS tags contradictory to the input query in existing reverse dictionary models.

We simply pass the sentence vector of the input query $\mathbf{v}$ to a single-layer perceptron:
\begin{equation}
\begin{aligned}
    \mathbf{sc}_{pos} = \mathbf{W}_{pos}   \mathbf{v}+\mathbf{b}_{pos},\\
\end{aligned}
\end{equation}
where $\mathbf{sc}_{pos}\in \mathbb{R}^{|\mathbb{P}|}$ records the prediction score of each POS tag, $\mathbf{W}_{pos} \in \mathbb{R}^{|\mathbb{P}|\times 2l}$ is a weight matrix, and $\mathbf{b}_{pos} \in \mathbb{R}^{|\mathbb{P}|}$ is a bias vector.

The confidence score of $w$ from the POS tag channel is the sum of the prediction scores of $w$'s POS tags:
\begin{equation}
sc_{w,pos} = \sum_{p \in \mathbb{P}_w} [\mathbf{sc}_{pos}]_{{\rm index}^{pos}(p)},
\end{equation}
where $[\mathbf{x}]_i$ denotes the $i$-th element of $\mathbf{x}$, and ${\rm index}^{pos}(p)$ returns the POS tag index of $p$.

\subsection{Internal Channel: Morpheme Predictor}
Most words are complex words consisting of more than one morphemes. 
We find there exists a kind of local semantic correspondence between the morphemes of a word and its definition or description.
For instance, the word ``expressway'' has two morphemes ``express'' and ``way'' and its dictionary definition is ``a wide \textbf{road} in a city on which cars can travel very \textbf{quickly}''. 
We can observe that the two words ``road'' and ``quickly'' semantically correspond to the two morphemes ``way'' and ``express'' respectively.
By predicting morphemes of the target word from the input query, a reverse dictionary can capture compositional information of the target word, which is complementary to contextual information of word embeddings.

We design a special morpheme predictor.
Different from the POS tag predictor, we allow each hidden state to be involved in morpheme prediction directly, and do max-pooling to obtain final morpheme prediction scores.
Specifically, we feed each non-directional hidden state to a single-layer perceptron and obtain local morpheme prediction scores:
\begin{equation}
\setlength{\abovedisplayskip}{3pt}
\setlength{\abovedisplayshortskip}{1pt}
\setlength{\belowdisplayskip}{3pt}
\setlength{\belowdisplayshortskip}{3pt}
    \begin{aligned}
        \mathbf{sc}^i_{mor} = \mathbf{W}_{mor}   \mathbf{h}_i+\mathbf{b}_{mor},
    \end{aligned}
\end{equation}
where $\mathbf{sc}^i_{mor} \in \mathbb{R}^{|\mathbb{M}|}$ measures the semantic correspondence between $i$-th word in the input query and each morpheme, $\mathbf{W}_{mor} \in \mathbb{R}^{\mathbb{M}| \times 2l}$ is a weight matrix, and $\mathbf{b}_{mor} \in \mathbb{R}^{|\mathbb{M}|} $ is a bias vector.
Then we do max-pooling over all the local morpheme prediction scores to obtain global morpheme prediction scores:
\begin{equation}
\setlength{\abovedisplayskip}{1pt}
\setlength{\abovedisplayshortskip}{0pt}
\setlength{\belowdisplayskip}{3pt}
\setlength{\belowdisplayshortskip}{3pt}
    \left[\mathbf{sc}_{mor} \right]_j=\max_{1\le i\le |Q|}[\mathbf{sc}^i_{mor}]_j.
\end{equation}

And the confidence score of $w$ from the morpheme channel is:
\begin{equation}
\setlength{\abovedisplayskip}{1pt}
\setlength{\abovedisplayshortskip}{0pt}
\setlength{\belowdisplayskip}{2pt}
\setlength{\belowdisplayshortskip}{2pt}
    sc_{w,mor}=\sum_{m \in \mathbb{M}_w} [\mathbf{sc}_{mor}]_{{\rm index}^{mor}(m)},
\end{equation}
where ${\rm index}^{mor}(m)$ returns the morpheme index of $m$.




\subsection{External Channel: Word Category Predictor}
Semantically related words often belong to different categories, although they have close word embeddings, e.g., ``car'' and ``road''.
Word category information is helpful in eliminating semantically related but not similar words from the results of reverse dictionaries \citep{MorinagaY18}.
There are many available word taxonomy systems which can provide hierarchical word category information, e.g., WordNet \cite{Miller1995WordNet}.
Some of them provides POS tag information as well, in which case POS tag predictor can be removed.


We design a hierarchical predictor to calculate prediction scores of word categories.
Specifically, each word belongs to a certain category in each layer of word hierarchy. 
We first compute the word category prediction score of each layer:
\begin{equation}
\setlength{\abovedisplayskip}{3pt}
\setlength{\abovedisplayshortskip}{3pt}
\setlength{\belowdisplayskip}{3pt}
\setlength{\belowdisplayshortskip}{3pt}
\begin{aligned}
    \mathbf{sc}_{cat,k} = \mathbf{W}_{cat,k}   \mathbf{v}+\mathbf{b}_{cat,k}, \\
\end{aligned}
\end{equation}
where $\mathbf{sc}_{cat,k} \in \mathbb{R}^{c_k}$ is the word category prediction score distribution of $k$-th layer, 
$\mathbf{W}_{cat,k} \in \mathbb{R}^{c_k \times 2l}$ is a weight matrix, $\mathbf{b}_{cat,k} \in \mathbb{R}^{c_k}$ is a bias vector, and $c_k$ is the category number of $k$-th layer.
Then the final confidence score of $w$ from the word category channel is the weighted sum of its  category prediction scores of all the layers:
\begin{equation}
\setlength{\abovedisplayskip}{2pt}
\setlength{\abovedisplayshortskip}{3pt}
\setlength{\belowdisplayskip}{2pt}
\setlength{\belowdisplayshortskip}{3pt}
\begin{aligned}
    sc_{w,cat} = \sum_{k=1}^{K} \beta_k [\mathbf{sc}_{cat,k}]_{{\rm index}_k^{cat}(w)},
\end{aligned}
\end{equation}
where $K$ is the total layer number of the word hierarchy, $\beta_k$ is a hyper-parameter controlling the relative weights, and ${\rm index}_k^{cat}(w)$ returns the category index of $w$ in the $k$-th layer.

\subsection{External Channel: Sememe Predictor}
In linguistics, a sememe is the minimum semantic unit of natural languages \citep{bloomfield1926set}.
Sememes of a word can accurately depict the meaning of the word. 
HowNet \citep{dong2003hownet} is the most famous sememe knowledge base.
It defines about $2,000$ sememes and uses them to annotate more than $100,000$ Chinese and English words by hand.
HowNet and its sememe knowledge has been widely applied to various NLP tasks including sentiment analysis \citep{fu2013multi}, word representation learning \citep{niu2017improved}, semantic composition \citep{qi2019modeling}, sequence modeling \citep{qin2019enhancing} and textual adversarial attack \citep{zang2019open}.

Sememe annotation of a word in HowNet includes hierarchical sememe structures as well as relations between sememes.
For simplicity, we extract a set of unstructured sememes for each word, in which case sememes of a word can be regarded as multiple semantic labels of the word.
We find there also exists local semantic correspondence between the sememes of a word and its description.
Still taking ``expressway'' as an example, its annotated sememes in HowNet are \texttt{route} and \texttt{fast}, which semantically correspond to the words in its definition ``road'' and ``quickly'' respectively.

Therefore, we design a sememe predictor similar to the morpheme predictor. 
Formally, we use $\mathbb{S}$ to represent the set of all sememes.
The sememe set of a word $w$ is $\mathbb{S}_w = \{ s_1, \cdots, s_{|\mathbb{S}_w|}\}$.
We pass each hidden state to a single-layer perceptron to calculate local sememe prediction scores:
\begin{equation}
    \begin{aligned}
        \mathbf{sc}^i_{sem} = \mathbf{W}_{sem}   \mathbf{h}_i+\mathbf{b}_{sem},
    \end{aligned}
\end{equation}
where $\mathbf{sc}^i_{sem} \in \mathbb{R}^{|\mathbb{S}|}$ indicates how corresponding between $i$-th word in the input query and each sememe, $\mathbf{W}_{sem} \in \mathbb{R}^{|\mathbb{S}| \times 2l}$ is a weight matrix, and $\mathbf{b}_{sem}$ is a bias vector.
Final sememe prediction scores are computed by doing max-pooling:
\begin{equation}
    \left[\mathbf{sc}_{sem} \right]_j=\max_{1\le i\le |Q|}[\mathbf{sc}^i_{sem}]_j.
\end{equation}

The confidence score of $w$ from the sememe channel is:
\begin{equation}
    sc_{w,sem}=\sum_{s \in \mathbb{S}_w} [\mathbf{sc}_{sem}]_{{\rm index}^{sem}(s)},
\end{equation}
where ${\rm index}^{sem}(s)$ returns the sememe index of $s$.

\subsection{Multi-channel Reverse Dictionary Model}
By combining the confidence scores of direct word prediction and indirect characteristic prediction, we obtain the final confidence score of a given word $w$ in our multi-channel reverse dictionary model:
\begin{equation}
\begin{aligned}
    sc_{w}=\lambda_{word}sc_{w,word}+\sum_{c\in \mathbb{C}}\lambda_{c}sc_{w,c},
\end{aligned}
\end{equation}
where $\mathbb{C}=\{pos,mor,cat,sem\}$ is the channel set, and $\lambda_{word}$ and $\lambda_{c}$ are the hyper-parameters controlling relative weights of corresponding terms.

As for training loss, we simply adopt the one-versus-all cross-entropy loss inspired by the sentence classification models.

\begin{table*}[!htpb]
    \centering
    \begin{tabular}{c|ccc|ccc|ccc}
    \toprule
        Model  & \multicolumn{3}{c|}{\textbf{Seen} Definition}  & \multicolumn{3}{c|}{\textbf{Unseen} Definition }  & \multicolumn{3}{c}{\textbf{Description}} \\ \hline
        OneLook & \textbf{0} & .66/.94/.95 & 200 & - & - & - & 5.5 & \textbf{.33}/.54/.76 & 332 \\ \hline
        BOW & 172 & .03/.16/.43 & 414 & 248 & .03/.13/.39 & 424 & 22 & .13/.41/.69 & 308 \\
        RNN & 134 & .03/.16/.44 & 375 & 171 & .03/.15/.42 & 404 & 17 & .14/.40/.73 & 274 \\
        RDWECI & 121 & .06/.20/.44 & 420 & 170 & .05/.19/.43 & 420 & 16 & .14/.41/.74 & 306 \\
        SuperSense & 378 & .03/.15/.36 & 462 & 465 & .02/.11/.31 & 454 & 115 & .03/.15/.47 & 396 \\
        MS-LSTM & \textbf{0} & \textbf{.92}/\textbf{.98}/\textbf{.99} & \textbf{65} & 276 & .03/.14/.37 & 426 & 1000 & .01/.04/.18 & 404 \\
        \hline
        BiLSTM & 25 & .18/.39/.63 & 363 & 101 & .07/.24/.49 & 401 & 5 & .25/.60/.83 & 214 \\
        +Mor & 24 & .19/.41/.63 & 345 & 80 & .08/.26/.52 & 399 & 4 & .26/.62/.85 & \textbf{198} \\
        +Cat & 19 & .19/.42/.68 & 309 & 68 & .08/.28/.54 & 362 & 4 & .30/.62/.85 & 206 \\
        +Sem & 19 & .19/.43/.66 & 349 & 80 & .08/.26/.53 & 393 & 4 & .30/\textbf{.64}/.87 & 218 \\ \hline
        Multi-channel & 16 & .20/.44/.71 & 310 & \textbf{54} & \textbf{.09}/\textbf{.29}/\textbf{.58} & \textbf{358} & \textbf{2} & .32/\textbf{.64}\textbf{/.88} & 203 \\
        \bottomrule 
        \multicolumn{3}{l|}{} & \multicolumn{2}{c}{\textit{median rank}} & \multicolumn{3}{r}{\textit{accuracy@1/10/100}} & \multicolumn{2}{c|}{\textit{rank variance}} \\
    \end{tabular}
    \caption{Overall reverse dictionary performance of all the models. }
    \label{tab:main-results-En}
\end{table*}

\section{Experiments}
In this section, we evaluate the performance of our multi-channel reverse dictionary model. 
We also conduct detailed quantitative analyses as well as a case study to explore the influencing factors in the reverse dictionary task and demonstrate the strength and weakness of our model.
We carry out experiments on both English and Chinese datasets. 
But due to limited space, we present our experiments on the Chinese dataset in the appendix.

\subsection{Dataset}
We use the English dictionary definition dataset created by \citet{Hill2016LearningTU}\footnote{The definitions are extracted from five electronic resources: \textit{WordNet}, \textit{The American Heritage Dictionary}, \textit{The Collaborative International Dictionary of English}, \textit{Wiktionary} and \textit{Webster's}.} as the training set. It contains about $100,000$ words and $900,000$ word-definition pairs. 
We have three test sets including: 
(1) \textbf{seen} definition set, which contains $500$ pairs of words and WordNet definitions existing in the training set and is used to assess the ability to recall previously encoded information; 
(2) \textbf{unseen} definition set, which also contains $500$ pairs of words and WordNet definitions but the words together with all their definitions have been excluded from the training set; 
and (3) \textbf{description set}, which consists of $200$ pairs of words and human-written descriptions and is a benchmark dataset created by \citet{Hill2016LearningTU} too.

To obtain the morpheme information our model needs, we use Morfessor \citep{Virpioja2013Morfessor} to segment all the words into morphemes.
As for the word category information, we use the \textit{lexical names} from WordNet \citep{Miller1995WordNet}. 
There are $45$ lexical names and the total layer number of the word category hierarchy is $1$.
Since the lexical names have included POS tags, e.g., \textit{noun.animal}, we remove the POS tag predictor from our model.
We use HowNet as the source of sememes. 
It contains $43,321$ English words manually annotated with $2,148$ different sememes in total.
We employ OpenHowNet \citep{qi2019openhownet}, the open data accessing API of HowNet, to obtain sememes of words.

\subsection{Experimental Settings}
\paragraph{Baseline Methods}
We choose the following models as the baseline methods: 
(1) OneLook, the most popular commercial reverse dictionary system, whose 2.0 version is used; 
(2) BOW and RNN with rank loss \citep{Hill2016LearningTU}, both of which are NLM based and the former uses a bag-of-words model while the latter uses an LSTM; 
(3) RDWECI \citep{MorinagaY18}, which incorporates category inference and is an improved version of BOW; 
(4) SuperSense \citep{pilehvar2019importance}, an improved version of BOW which uses pretrained sense embeddings to substitute target word embeddings;
(5) MS-LSTM \citep{kartsaklis2018mapping}, an improved version of RNN which uses graph-based WordNet synset embeddings together with a multi-sense LSTM to predict \textit{synsets} from descriptions and claims to produce state-of-the-art performance;
and (6) BiLSTM, the basic framework of our multi-channel model.

\paragraph{Hyper-parameters and Training}
For our model, the dimension of non-directional hidden states is $300\times 2$, the weights of different channels are equally set to $1$, and the dropout rate is 0.5.
For all the models except MS-LSTM, we use the 300-dimensional word embeddings pretrained on GoogleNews with \texttt{word2vec}\footnote{\url{https://code.google.com/archive/p/word2vec/}}, and the word embeddings are fixed during training.
For all the other baseline methods, we use their recommended hyper-parameters.
For training, we adopt Adam as the optimizer with initial learning rate $0.001$, and the batch size is $128$. 









\paragraph{Evaluation Metrics}
Following previous work, we use three evaluation metrics: the median rank of target words (lower better), the accuracy that target words appear in top 1/10/100 (acc@1/10/100, higher better) and the standard deviation of target words' ranks (rank variance, lower better).
Notice that MS-LSTM can only predict WordNet synsets. Thus, we map the target words to corresponding WordNet synsets (target synsets) and calculate the accuracy and rank variance of the target synsets.

\begin{table*}
\centering
\begin{tabular}{c|ccc|ccc|ccc}
\toprule
Prior Knowlege & \multicolumn{3}{c|}{\textbf{Seen} Definition} & \multicolumn{3}{c|}{\textbf{Unseen} Definition} & \multicolumn{3}{c}{\textbf{Description}} \\ 
\hline
None & 16 & .20/.44/.71 & 310 & 54 & .09/.29/.58 & 358 & 2.5 & .32/.64/.88 & 203 \\
POS Tag & 13 & .21/.45/.72 & 290 & 45 & .10/.31/.60 & 348 & 3 & .35/.65/.91 & 174 \\
Initial Letter & 1 & .39/.73/.90 & 270 & 4 & .26/.63/.85 & 348 & 0 & .62/.90/.97 & 160 \\
Word Length & 1 & .40/.71/.90 & 269 & 6 & .25/.56/.84 & 346 & 0 & .55/.85/.95 & 163 \\
 \bottomrule
\multicolumn{3}{l|}{} & \multicolumn{2}{c}{\textit{median rank}} & \multicolumn{3}{r}{\textit{accuracy@1/10/100}} & \multicolumn{2}{c|}{\textit{rank variance}} \\
\end{tabular}
\caption{Reverse dictionary performance with prior knowledge.}
\label{tab:result-priori-En}
\end{table*}

\subsection{Overall Experimental Results}

Table \ref{tab:main-results-En} exhibits reverse dictionary performance of all the models on the three test sets, where ``Mor'', ``Cat'' and ``Sem'' represent the morpheme, word category and sememe predictors respectively.
Notice that the performance of OneLook on the unseen test set is meaningless because we cannot exclude any definitions from its definition bank, hence we do not list corresponding results.
From the table, we can see:

(1) Compared with all the baseline methods other than OneLook, our multi-channel model achieves substantially better performance on the unseen definition set and the description set, which verifies the absolute superiority of our model in generalizing to the novel and unseen input queries.

(2) OneLook significantly outperforms our model when the input queries are dictionary definitions.
This result is expected because the input dictionary definitions are already stored in the database of OneLook and even simple text matching can easily handle this situation.
However, the input queries of a reverse dictionary cannot be exact dictionary definitions in reality. 
On the description test set, our multi-channel model achieves better overall performance than OneLook. 
Although OneLook yields slightly higher acc@1, it has limited value in terms of practical application, because people always need to pick the proper word from several candidates, not to mention the fact that the acc@1 of OneLook is only $0.33$.

(3) MS-LSTM performs very well on the seen definition set but badly on the description set, which manifests its limited generalization ability and practical value.
Notice that when testing MS-LSTM, the searching space is the whole synset list rather than the synset list of the test set, which causes the difference in performance on the unseen definition set measured by us and recorded in the original work \citep{kartsaklis2018mapping}.

(4) All the BiLSTM variants enhanced with different information channels (+Mor, +Cat and +Sem) perform better than vanilla BiLSTM. 
These results prove the effectiveness of predicting characteristics of target words in the reverse dictionary task.
Moreover, our multi-channel model achieves further performance enhancement as compared with the single-channel models, which demonstrates the potency of characteristic fusion and also verifies the efficacy of our multi-channel model.

(5) BOW performs better than RNN, which is consistent with the findings from \citet{Hill2016LearningTU}. 
However, BiLSTM far surpasses BOW as well as RNN. 
This verifies the necessity for bi-directional encoding in RNN models, and also shows the potential of RNNs.



\subsection{Performance with Prior Knowledge}
In practical application of reverse dictionaries, extra information about target words in addition to descriptions may be known.
For example, we may remember the initial letter of the word we forget, or the length of the target word is known in crossword game.
In this subsection, we evaluate the performance of our model with the prior knowledge of target words, including POS tag, initial letter and word length. 
More specifically, we extract the words satisfying given prior knowledge from the top $1,000$ results of our model, and then reevaluate the performance.
The results are shown in Table \ref{tab:result-priori-En}.

We can find that any prior knowledge improves the performance of our model to a greater or lesser extent, which is an expected result.
However, the performance boost brought by the initial letter and word length information is much bigger than that brought by the POS tag information.
The possible reasons are as follows. 
For the POS tag, it has been already predicted in our multi-channel model, 
hence the improvement it brings is limited, which also demonstrates that our model can do well in POS tag prediction.
For the initial letter and word length, they are hard to predict according to a definition or description and not considered in our model. Therefore, they can filter many candidates out and markedly increase performance.


\subsection{Analyses of Influencing Factors}
In this subsection, we conduct quantitative analyses of the influencing factors in reverse dictionary performance. 
To make results more accurate, we use a larger test set consisting of $38,113$ words and $80,658$ \textbf{seen} pairs of words and WordNet definitions.
Since we are interested in the features of target words, we exclude MS-LSTM that predicts WordNet synsets.


\begin{figure}[!t]
    \centering
    \includegraphics[width=.99\columnwidth]{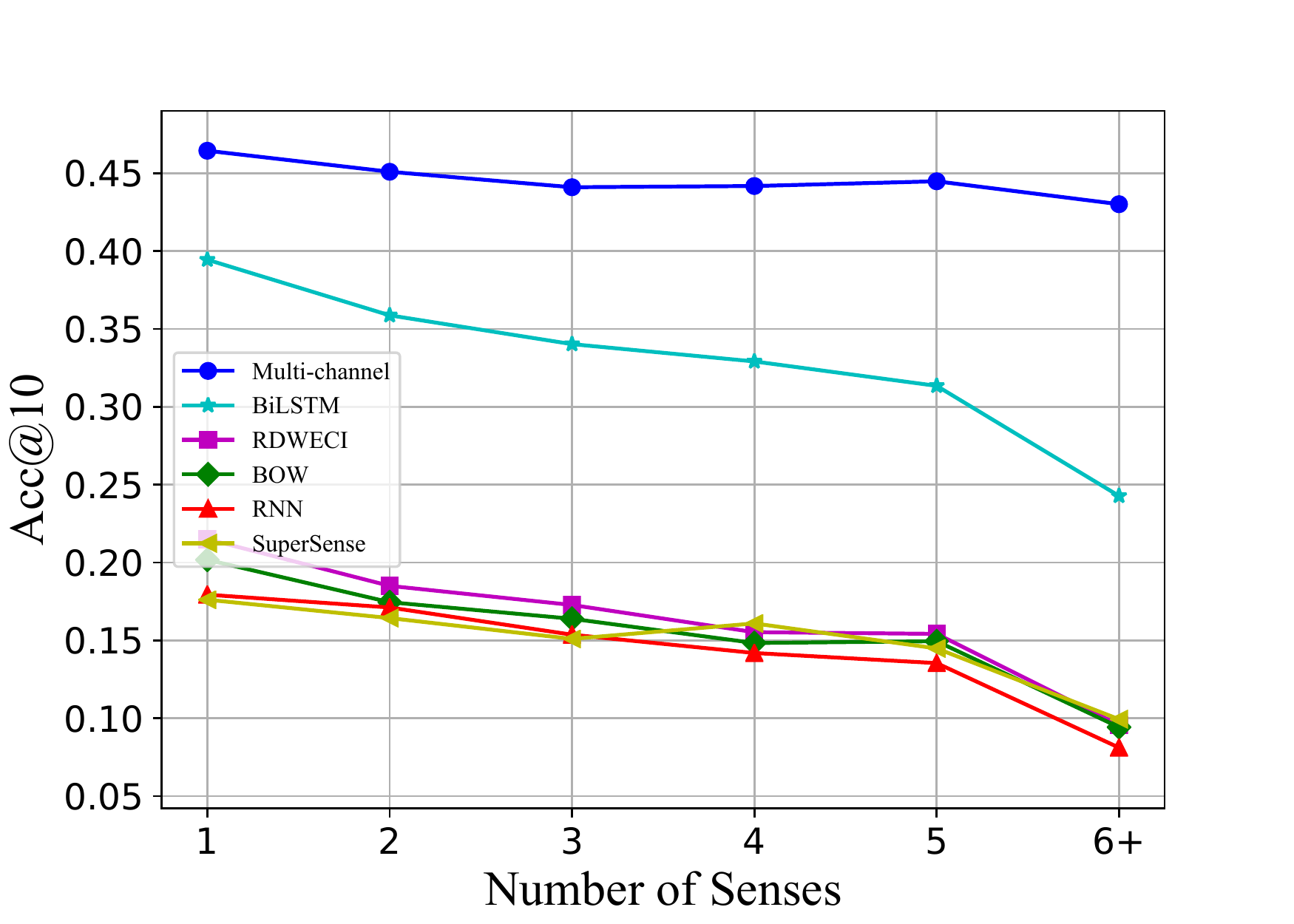}
    \caption{Acc@10 on words with different sense numbers. The numbers of words are $21,582$, $8,266$, $3,538$, $1,691$, $953$ and $2,083$ respectively.}
    \label{fig:defi-num}
\end{figure}

\paragraph{Sense Number}
Figure \ref{fig:defi-num} exhibits the acc@10 of all the models on the words with different numbers of senses.
It is obvious that performance of all the models declines with the increase in the sense number, which indicates that polysemy is a difficulty in the task of reserve dictionary. 
But our model displays outstanding robustness and its performance hardly deteriorates even on the words with the most senses.


\begin{figure}[!t]
    \centering
    \includegraphics[width=.99\columnwidth]{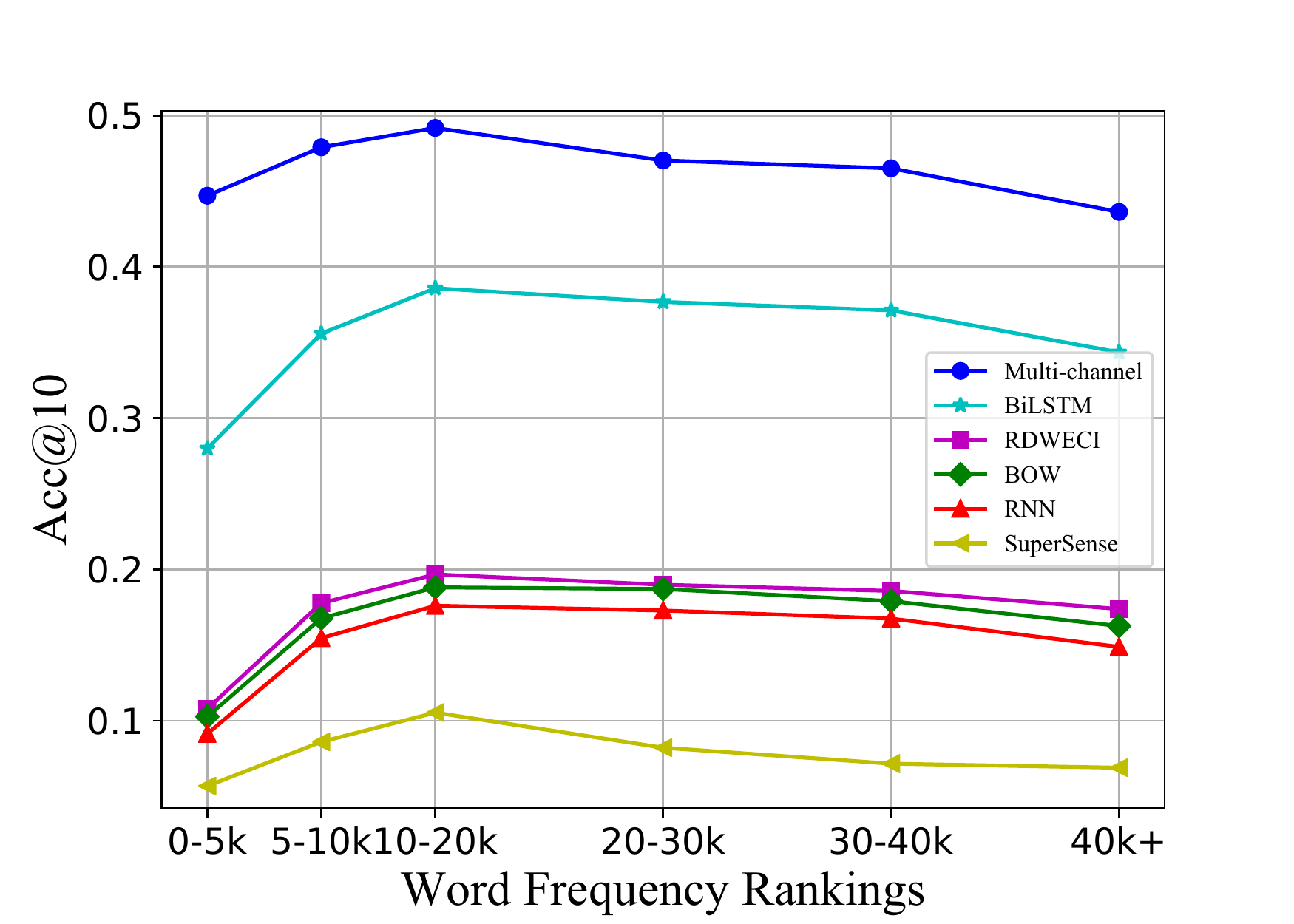}
    \caption{Acc@10 on ranges of different word frequency rankings. The number of words in each range is $3,299$, $3,243$, $3,515$, $3,300$, $5,565$ and $19,191$ respectively.}
    \label{fig:word-freq}
\end{figure}

\paragraph{Word Frequency}
Figure \ref{fig:word-freq} displays all the models' performance on the words within different ranges of word frequency ranking. 
We can find that the most frequent and infrequent words are harder to predict for all the reverse dictionary models. 
The most infrequent words usually have poor embeddings, which may damage the performance of NLM based models.
For the most frequent words, on the other hand, although their embeddings are better, they usually have more senses. 
We count the average sense numbers of all the ranges, which are $5.6$, $3.2$, $2.6$, $2.1$, $1.7$ and $1.4$ respectively. 
The first range has a much larger average sense number, which explains its bad performance.
Moreover, our model also demonstrates remarkable robustness.



\begin{figure}[!t]
    \centering
    \includegraphics[width=.99\columnwidth]{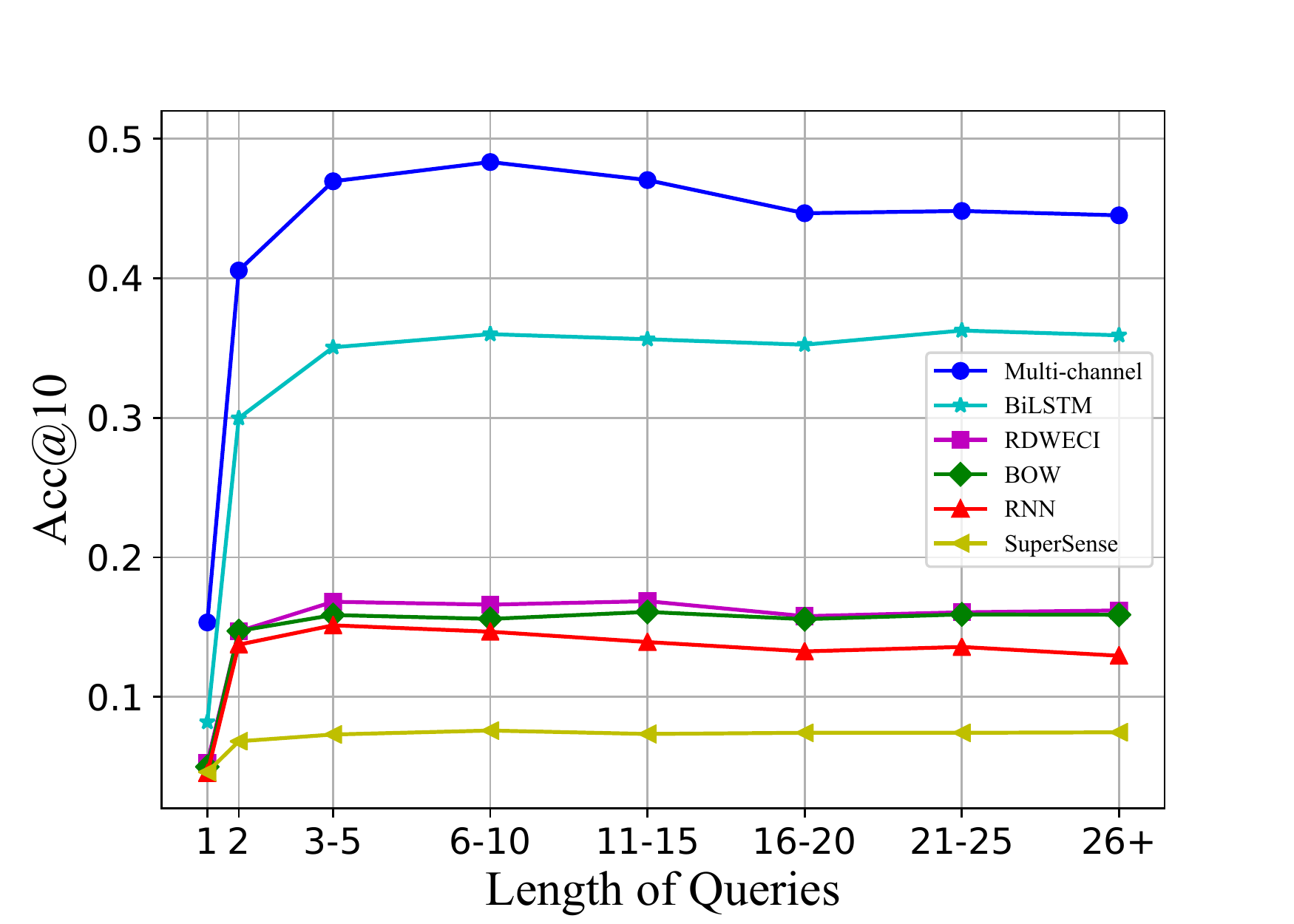}
    \caption{Acc@10 on ranges of different query length. The number of queries in each range is $672$, $3,609$, $21,113$, $31,013$, $14,684$, $5,803$, $2,316$ and $1,448$ respectively.}
    \label{fig:defi-len}
\end{figure}

\paragraph{Query Length} 
The effect of query length on reverse dictionary performance is illustrated in Figure \ref{fig:defi-len}. 
When the input query has only one word, the system performance is strikingly poor, especially our multi-channel model.
This is easy to explain because the information extracted from the input query is too limited.
In this case, outputting the synonyms of the query word is likely to be a better choice.


\subsection{Case study}
In this subsection, we give two cases in Table \ref{tab:case-study} to display the strength and weakness of our reverse dictionary model.
For the first word ``postnuptial'', our model correctly predicts its morpheme ``post'' and sememe ``GetMarried'' from the words ``after'' and ``marriage'' in the input query. 
Therefore, our model easily finds the correct answer.
For the second case, the input query describes a rare sense of the word ``takeaway''. 
HowNet has no sememe annotation for this sense, and morphemes of the word are not semantically related to any words in the query either. 
Our model cannot solve this kind of cases, which is in fact hard to handle for all the NLM based models. 
In this situation, the text matching methods, which return the words whose stored definitions are most similar to the input query, may help.




\begin{table}[!t]
\centering
\resizebox{.99\columnwidth}{!}{
    \begin{tabular}{l|l}
    \toprule
    Word & Input Query \\ \hline
    \bf{postnuptial} & relating to events after a marriage.\\ \hline
    \bf{takeaway} & concession made by a labor union to a company. \\
    \bottomrule
    \end{tabular}
}
\caption{Two reverse dictionary cases.}
\label{tab:case-study}
\end{table}

\section{Conclusion and Future Work}

In this paper, we propose a multi-channel reverse dictionary model, which incorporates multiple predictors to predict characteristics of target words from given input queries. 
Experimental results and analyses show that our model achieves the state-of-the-art performance and also possesses outstanding robustness.

In the future, we will try to combine our model with text matching methods to better tackle extreme cases, e.g., single-word input query.
In addition, we are considering extending our model to the cross-lingual reverse dictionary task. 
Moreover, we will explore the feasibility of transferring our model to related tasks such as question answering.

\section{Acknowledgements}
This work is funded by the Natural Science Foundation of China (NSFC) and the German Research Foundation (DFG) in Project Crossmodal Learning, NSFC 61621136008 / DFG TRR-169.
Furthermore, we thank the anonymous reviewers
for their valuable comments and suggestions.

\fontsize{9pt}{10pt} \selectfont
\bibliographystyle{aaai} 
\bibliography{AAAI-ZhangL.3674} 

\newpage
\appendix

\section{Appendix: Experiments on Chinese Dataset}

In this section, we evaluate our multi-channel reverse dictionary model on the Chinese dataset.

\subsection{Dataset}
For Chinese, we build a dictionary definition dataset as the training set. It contains $69,000$ words and $84,694$ word-definition pairs, and the definitions are extracted from \textit{Modern Chinese Dictionary (6th Edition)}\footnote{\url{http://www.cp.com.cn/book/978-7-100-08467-3 44.html}}, an authoritative Chinese dictionary. 
There are four test sets including (1) \textbf{seen} and \textbf{unseen} definition sets of size $2,000$, which are built in the similar way to English; (2) \textbf{description set}, which is created by us and contains $200$ word-description pairs given by Chinese native speakers; and (3) \textbf{question set}, which collects $272$ real-world Chinese exam question-answers of writing the right word given a description from the Internet.

For the morpheme information, we simply cut each word into Chinese characters as morphemes.
As for the word category information, we use HIT-IR Tongyici Cilin\footnote{\url{https://github.com/yaleimeng/Final_word_Similarity/tree/master/cilin}}. 
It has five levels of word category hierarchy, and we only use the first four levels. 
The numbers of categories in each level are $12$, $95$, $1,425$ and $4,097$, respectively. 
We also use HowNet as the source of sememes in the same way as English.
The POS tags can be extracted from Modern Chinese Dictionary (6th Edition). We use all its $12$ POS tags.

\subsection{Experimental Settings}
\paragraph{Baseline Methods}
We choose the same baseline methods as English except OneLook, SuperSense and MS-LSTM. 
We exclude OneLook because it only supports English reverse dictionary search and there are no Chinese reverse dictionary systems.
In addition, SuperSense and MS-LSTM rely on WordNet but the Chinese version of WordNet contains too few words. 
So we do not make comparison with them, either.
More specifically, the baselines are 
(1) BOW and RNN with rank loss \citep{Hill2016LearningTU}, both of which are NLM based and the former uses a bag-of-words model while the latter uses an LSTM;
(2) RDWECI \citep{MorinagaY18}, which incorporates category inference and is an improved version of BOW; 
and (3) BiLSTM, the basic framework of our multi-channel model.

\paragraph{Hyper-parameters and Training}
For our model on the Chinese dataset, the dimension of non-directional hidden states is $200\times 2$, which is different from the model of English. 
The weights of different channels are equally set to $1$.
For the baseline methods, we use their recommended hyper-parameters.
For all the models, we use the 200-dimensional word embeddings pretrained on the SogouT corpus\footnote{\url{https://www.sogou.com/labs/resource/t.php}} with \texttt{word2vec}\footnote{\url{https://code.google.com/archive/p/word2vec/}}, and the word embeddings are fixed during training.
For training, we adopt Adam as the optimizer with initial learning rate $0.001$, and the batch size is $128$, which are all the same as that of English experiments.

\paragraph{Evaluation Protocols}
Same as the English experiments, we utilize three metrics including (1) the median rank of the target words; (2) the accuracy that the target words appears in top 1/10/100; and (3) the standard deviation of the target words' ranks.

\subsection{Overall Experimental Results}

Table \ref{tab:main-results-Ch} exhibits reverse dictionary performance of all the models on the four test sets, where ``+POS'',``+Mor'', ``+Cat'' and ``+Sem'' represent the POS tag, morpheme, word category and sememe predictors respectively.
From the table, we can see:

\begin{table*}[!t]
\small
\centering
\begin{tabular}{c|ccc|ccc|ccc|ccc}
\toprule
\multicolumn{1}{c|}{Model} & \multicolumn{3}{c|}{\textbf{Seen} Definition} & \multicolumn{3}{c|}{\textbf{Useen} Definition} & \multicolumn{3}{c|}{\textbf{Description}} & \multicolumn{3}{c}{\textbf{Question}} \\ \hline
BOW & 59 & .08/.28/.56 & 403 & 65 & .08/.28/.53 & 411 & 40 & .07/.30/.60 & 357 & 42 & .10/.28/.63 & 362 \\
RNN & 69 & .05/.23/.55 & 379 & 103 & .05/.21/.49 & 405 & 79 & .04/.26/.53 & 361 & 56 & .07/.27/.60 & 346 \\
RDWECI & 56 & .09/.31/.56 & 423 & 83 & .08/.28/.52 & 436 & 32 & .09/.32/.59 & 376 & 45 & .12/.32/.61 & 384 \\ \hline
BiLSTM & 4 & .28/.58/.78 & 302 & 14 & .15/.45/.71 & 343 & 13 & .14/.44/.78 & \textbf{233} & 4 & .30/.61/.82 & 243 \\
+POS & 4 & .28/.58/.78 & 309 & 14 & .16/.45/.71 & 346 & 13 & .14/.44/.79 & 255 & 5 & .25/.59/.79 & 271 \\
+Mor & 1 & .43/.73/.87 & 260 & 11 & \textbf{.19}/.47/.73 & 332 & 8 & .22/.52/\textbf{.83} & 251 & 1 & .42/.73/.86 & 227 \\
+Cat & 4 & .29/.58/.78 & 319 & 16 & .14/.43/.70 & 356 & 13 & .16/.45/.77 & 289 & 3 & .33/.62/.82 & 246 \\
+Sem & 4 & .29/.60/.80 & 298 & 14 & .16/.45/.72 & 340 & 12 & .15/.45/.75 & 244 & 4 & .34/.61/.83 & 231 \\ \hline
Multi-channel & \textbf{1} & \textbf{.49}/\textbf{.78}/\textbf{.90} & \textbf{220} & \textbf{10} & .18/\textbf{.49}/\textbf{.76} & \textbf{310} & \textbf{5} & \textbf{.24}/\textbf{.56}/.82 & 260 & \textbf{0} & \textbf{.50}/\textbf{.73}/\textbf{.90} & \textbf{223} \\
\bottomrule
\multicolumn{6}{l|}{} & \multicolumn{2}{c}{\textit{median rank}} & \multicolumn{3}{r}{\textit{accuracy@1/10/100}} & \multicolumn{2}{c|}{\textit{rank variance}} \\
\end{tabular}
\caption{Overall reverse dictionary performance of all the models. }
\label{tab:main-results-Ch}
\end{table*}

\begin{table*}[!t]
\small
\centering
\begin{tabular}{c|ccc|ccc|ccc|ccc}
\toprule
Prior Knowledge & \multicolumn{3}{c|}{\textbf{Seen} Definition} & \multicolumn{3}{c|}{\textbf{Useen} Definition} & \multicolumn{3}{c|}{\textbf{Description}} & \multicolumn{3}{c}{\textbf{Question}} \\ \hline
None & 1 & .49/.78/.90 & 220 & 10 & .18/.49/.76 & 310 & 5 & .24/.56/.82 & 260 & 0 & .50/.73/.90 & 223 \\
POS Tag & 1 & .50/.79/.90 & 222 & 9 & .18/.51/.77 & 307 & 4 & .24/.61/.85 & 252 & 0 & .50/.74/.90 & 223 \\
Initial Char & 0 & .74/.89/.92 & 220 & 0 & .55/.82/.86 & 304 & 0 & .61/.88/.93 & 239 & 0 & .84/.95/.95 & 213 \\
Word Length & 0 & .54/.82/.91 & 217 & 6 & .23/.57/.81 & 297 & 3 & .32/.68/88 & 242 & 0 & .62/.85/.94 & 212 \\ 
\bottomrule
\multicolumn{6}{l|}{} & \multicolumn{2}{c}{\textit{median rank}} & \multicolumn{3}{r}{\textit{accuracy@1/10/100}} & \multicolumn{2}{c|}{\textit{rank variance}} \\
\end{tabular}
\caption{Reverse dictionary performance with prior knowledge.}
\label{tab:result-priori-Ch}
\end{table*}

(1) Our multi-channel model achieves substantially better performance than all the baseline methods on all the four test sets, which demonstrates the superiorty of our model.
In addition, similar to the results of the English experiments, our model can also generalize well to the novel, unseen input queries.

(2) All the BiLSTM variants enhanced with different information channels (+POS, +Mor, +Cat and +Sem) perform better than vanilla BiLSTM except the evaluation for BiLSTM+POS on the Question test set. 
That is because words in the Question test set are all idioms and most of them have no POS tags.
Basically, the results prove the effectiveness of all the four information channels.

(3) BiLSTM's better performance than BOW and RNN demonstrate the necessity of bi-directional encoding in RNN models, although BOW performs also better than RNN here.

(4) The results on the Question test set show that our model is also good at question-answer exercise problems in real-world exams.

\subsection{Performance with Prior Knowledge}

Similar to the English experiments, we use the prior knowledge of the target word to evaluate the performance of our model on the Chinese dataset in the same way.

The results are shown in Table \ref{tab:result-priori-Ch}.
We can also find that any prior knowledge can improve our model's performance, especially the initial character information. 
That is presumably because the average character number of Chinese words is much less than that of English words and the search space is reduced to be smaller.
Similar to English, the performance improvement of our model given POS tag information is also insignificant, which also demonstrates that our model can do well in POS tag prediction.


\end{document}